\documentclass{article}

\usepackage{arxiv}

\usepackage[utf8]{inputenc} 
\usepackage[T1]{fontenc}    
\usepackage{hyperref}       
\usepackage{url}            
\usepackage{booktabs}       
\usepackage{amsfonts}       
\usepackage{nicefrac}       
\usepackage{microtype}      
\usepackage{lipsum}		
\usepackage{graphicx}
\usepackage{doi}

\usepackage{amsmath,amssymb,amsfonts,amsbsy}
\usepackage{algorithmic}
\usepackage{graphicx}
\usepackage{xcolor}
\usepackage{textcomp}
\usepackage{subfigure}
\usepackage{adjustbox}
\usepackage{comment}
\usepackage[]{algorithm2e}
\usepackage{float}

\usepackage{tikz}
\usetikzlibrary{shapes,arrows,shadows,positioning}
\usetikzlibrary{matrix,chains,positioning,decorations.pathreplacing}

\newcommand{\hlr}[1]{\textcolor{red}{#1}}
\newcommand{\hlb}[1]{\textcolor{blue}{#1}}

\title{Pipeline para la detección del trastorno específico del lenguaje (SLI) a partir de transcripciones de narrativas espontáneas}

\author{Santiago Arena,
  Antonio~Quintero-Rinc\'on, \\ Departamento de Ciencia de Datos, Laboratorio de ciencia de Datos e Inteligencia Artificial\\ Universidad Católica de Argentina (UCA) \\ Buenos Aires, Argentina\\
    \texttt{santiagoarena@uca.edu.ar, antonioquintero@uca.edu.ar}}

\renewcommand{\shorttitle}{\textit{arXiv} Template}

\hypersetup{
pdftitle={A template for the arxiv style},
pdfsubject={q-bio.NC, q-bio.QM},
pdfauthor={David S.~Hippocampus, Elias D.~Striatum},
pdfkeywords={First keyword, Second keyword, More},
}

\begin{document}
\maketitle
\begin{abstract}
El Trastorno Específico del Lenguaje (SLI) es un trastorno que afecta la comunicación y puede afectar tanto la comprensión como la expresión. Este estudio se centra en la detección eficaz del SLI en niños, empleando transcripciones de narrativas espontáneas tomadas en 1063 entrevistas. Para dicho fin, proponemos un \emph{pipeline} de tres etapas en cascada. En la primera etapa, se hace una extracción de características y una reducción de dimensionalidad de los datos usando en conjunto, los métodos de \emph{Random Forest} (RF) y correlación de \emph{Spearman}. En la segunda etapa se estiman las variables más predictivas de la primera etapa usando regresión logística, las cuales son usadas en la última etapa, para detectar el trastorno SLI en niños a partir de transcripciones de narrativas espontáneas usando un clasificador de vecinos más cercanos.
Los resultados revelaron una precisión del 97,13\% en la identificación del SLI, destacándose aspectos como el largo de las respuestas, la calidad de sus enunciados y la complejidad del lenguaje. Este nuevo enfoque enmarcado en procesamiento natural del lenguaje, ofrece beneficios significativos al campo de la detección de SLI, al evitar complejas variables subjetivas y centrarse en métricas cuantitativas directamente relacionadas con el desempeño del niño.

\end{abstract}

\section{Introducci\'on}
El Trastorno Espec\'ifico del Lenguaje (SLI por sus siglas en ingl\'es, \emph{Specific Language Impairment}), tambi\'en llamado disfasia infantil, afecta entre el $2$ y el $11$ porciento de los ni\~nos menores de $13$ a\~nos, siendo caracterizado por deficiencias en el lenguaje, sin discapacidad o irregularidades mentales o f\'isica evidente. En otras palabras, es un trastorno que afecta la comunicaci\'on y puede afectar tanto la comprensi\'on como la expresi\'on \cite{Laurence2014SLI}. Es m\'as com\'un en ni\~nos que en ni\~nas, con un fuerte v\'inculo gen\'etico, dado que entre el $50\%$ y el $70\%$ de los ni\~nos con SLI tienen un miembro de la familia con esta afecci\'on \cite{Sharma2022}. Un ni\~no con SLI a menudo tiene antecedentes de retraso en el desarrollo del lenguaje expresivo, presentando s\'intomas como dificultades en la construcci\'on de oraciones, desarrollo del vocabulario, entablar conversaciones, comprender y seguir reglas gramaticales y/o comprender instrucciones habladas \cite{Barua2023}.
Un trastorno del lenguaje (SLI) ocurre cuando un ni\~no muestra una incapacidad persistente para adquirir y usar habilidades ling\"u\'isticas, como se proyectar\'ia seg\'un las expectativas normativas basadas en la edad \cite{APA2022}; este trastorno se considera \emph{primario} o \emph{espec\'ifico} cuando no hay una explicaci\'on clara para estos retrasos en las habilidades ling\"u\'isticas. La mayor\'ia de los ni\~nos identificados con trastorno primario del lenguaje al ingresar a la escuela seguir\'an teniendo habilidades ling\"u\'isticas significativamente deprimidas con el tiempo \cite{Webster2004}, mostrar\'an dificultades con la preparaci\'on para el jard\'in de infantes \cite{Pentimonti2016}, y tendr\'an dificultades para aprender a leer \cite{Catts2002}, este \'ultimo debido en parte a sus efectos en habilidades ling\"u\'isticas de nivel superior\cite{Hogan2011}. El trastorno del lenguaje en la primera infancia tambi\'en est\'a vinculado a un mayor riesgo de preocupaciones psiqui\'atricas, dificultades de atenci\'on, problemas socio-conductuales y discapacidades de aprendizaje en la adolescencia \cite{beitchman1996seven}\cite{stanton2007social}. Dada la incidencia relativamente alta de esta discapacidad infantil y sus efectos significativos en numerosas \'areas de desarrollo, existe un gran inter\'es en garantizar una identificaci\'on precisa y una intervenci\'on temprana para los ni\~nos afectados en edad temprana \cite{justice2019identifying}.

En la literatura se destaca la importancia de detectar trastornos espec\'ificos del lenguaje en edades tempranas, bas\'andose en asociaciones de los pacientes con d\'eficits en diversos aspectos del lenguaje \cite{justice2019identifying}. As\'i mismo, se destaca la importancia de distinguir SLI de otros trastornos, como el autismo, para mejorar la precisi\'on diagn\'ostica \cite{Gabani2011ExploringAC}.
M\'ultiples enfoques se han desarrollado utilizando diversas herramientas de ML para la clasificaci\'on de SLI en ni\~nos. Mac Farlane et al. \cite{MacFarlane2017} utiliza un esquema de redes neuronales para clasificar ni\~nos a trav\'es de \'indices de desempe\~no de $10$ indicadores de confecci\'on personal, estos demostraron que es posible clasificar ni\~nos biling\"ues en los lenguajes de espa\~nol e ingl\'es. En el trabajo de Sharma et al. \cite{Sharma2022}, se propuso un modelo que aprovecha datos directos de las transcripciones sin ning\'un procesamiento, utilizando redes neuronales convolucionales y aprendizaje profundo para segregar los pacientes entre SLI y desarrollo t\'ipico (TD). Kaushik et al. \cite{Kaushik2021}  aplica un m\'etodo de detecci\'on de SLI denominado SLINet, donde a partir de una red neuronal convolucional 2D obtuvieron en $98$ sujetos ($54$ SLI y $44$ controles), una precisi\'on del $99.09\%$ utilizando validaci\'on cruzada de diez pliegues. Por otra parte, Gray et al. \cite{Gray2003} present\'o un enfoque para la detecci\'on de SLI en ni\~nos donde utilizaron la fiabilidad de prueba/reprueba y el par\'ametro de precisi\'on diagn\'ostica para evaluar el diagn\'ostico de SLI. Para este prop\'osito, se realiz\'o una evaluaci\'on en repetici\'on de no-palabras, serie de d\'igitos, bater\'ia de evaluaci\'on \emph{Kaufman} para Ni\~nos y el test de lenguaje expresivo fotogr\'afico estructurado, los autores informaron una especificidad y sensibilidad del $100\%$ y $95\%$, respectivamente con $44$ ni\~nos en edad preescolar ($22$ SLI, $22$ TD). Armon-Lotem y Meir \cite{ArmonLotem2016} propusieron un m\'etodo para la identificaci\'on de SLI en ni\~nos biling\"ues (hebreo, ruso), su estudio utiliz\'o pruebas de repetici\'on de d\'igitos, no-palabras y oraciones, logrando tasas de precisi\'on, sensibilidad y especificidad del $94\%$, $80\%$ y $97\%$, respectivamente, para ambos idiomas en $230$ ni\~nos mono y biling\"ues ($175$ TD, $55$ SLI). Slogrove y Haar \cite{slogrove2020} aplicaron coeficientes cepstrales de frecuencia \emph{Mel} de se\~nales de voz para la detecci\'on de SLI. Alcanzaron una tasa de precisi\'on del $99\% $utilizando un clasificador \emph{Random Forest} de forma aleatoria. Reddy et al. \cite{reddy2020} utilizaron caracter\'isticas de la fuente glotal, coeficientes cepstrales de frecuencia \emph{Mel} y una red neuronal feed-forward para la detecci\'on de SLI en ni\~nos. Su estudio utiliz\'o se\~nales de habla de $54$ pacientes con SLI y $44$ ni\~nos con TD. Informaron una precisi\'on del $98.82\%$ con selecci\'on de caracter\'isticas. Oliva et al. \cite{oliva2013} propusieron un m\'etodo de detecci\'on de SLI utilizando t\'ecnicas de aprendizaje autom\'atico. En su estudio, se utilizaron datos c\'omo la longitud media de las emisiones, oraciones no gramaticales, uso correcto de: art\'iculos, verbos, \emph{cl\'iticos},  argumentos tem\'aticos y proporci\'on de estructuras ditransitivas, entre otras, de 24 ni\~nos con SLI y 24 ni\~nos con TD, informaron tasas de sensibilidad y especificidad del $97\%$ y $100\%$, respectivamente. 
SLI actualmente es un campo de estudio de inter\'es en la comunidad cient\'ifica, como se ha expuesto en el estado-del-arte. La literatura resalta la importancia de la detecci\'on temprana de trastornos espec\'ificos del lenguaje, como el SLI, y la necesidad de distinguirlo de otros trastornos para un diagn\'ostico preciso. Varios estudios han explorado enfoques de aprendizaje autom\'atico, como redes neuronales y an\'alisis de caracter\'isticas de voz, con resultados prometedores en la clasificaci\'on de SLI versus desarrollo t\'ipico. Adem\'as, la detecci\'on de SLI es de sumo inter\'es en la automatizaci\'on del proceso mediante t\'ecnicas de Procesamiento Natural del Lenguaje (NLP) y \emph{Machine Learning} (ML) y en especial, en el dise\~no de instrumentos de medici\'on que se basen en aspectos cuantitativos del diagn\'ostico del paciente \cite{Sharma2022}. Precisamente, este estudio se centra en ni\~nos diagnosticados con SLI y busca identificar marcadores ling\"u\'isticos que permitan una detecci\'on temprana y eficiente de este trastorno. 

La presente investigaci\'on tiene como objetivo desarrollar un \emph{pipeline} en cascada usando cl\'asicas t\'ecnicas de ML, para detectar el trastorno SLI en ni\~nos a partir de transcripciones de narrativas espont\'aneas. Para ello, proponemos usar los m\'etodos de \emph{Random Forest} (RF) y correlaci\'on en conjunto, c\'omo selectores de caracter\'isticas y as\'i obtener una reducci\'on de dimensionalidad de los datos. Luego con estos datos, se usa el modelo de regresi\'on log\'istica, con el objetivo de obtener solamente las variables m\'as predictivas, para finalmente, usar un clasificador de vecinos m\'as cercano para detectar SLI.


El art\'iculo est\'a organizado de la siguiente manera. La Secci\'on \ref{sec:met} presenta la metodolog\'ia propuesta, donde se introduce el esquema del \emph{pipeline}, se explican conceptos clave como reducci\'on de dimensionalidad, los modelos aplicados en el mismo y las m\'etricas utilizadas para evaluar los resultados. Luego en la Secci\'on \ref{sec:res}, se detallan los resultados obtenidos en las distintas etapas del proceso. Se presentan los an\'alisis y las interpretaciones correspondientes a cada paso del m\'etodo propuesto, incluyendo la evaluaci\'on de la eficacia de las t\'ecnicas empleadas. Adem\'as, se discuten los hallazgos significativos y se comparan con resultados previos en la literatura, con el objetivo de validar y contextualizar los nuevos resultados obtenidos. Finalmente, en la Secci\'on \ref{sec:con} se extraen conclusiones, se realizan comparaciones,  se plantean limitaciones, fortalezas y se discute sobre trabajos futuros.
\section{Metodolog\'ia}
\label{sec:met}
El \emph{pipeline} en cascada propuesto, se compone de tres etapas, ver Figura~\ref{fig:bloques}. En la primera etapa (letras de color azul), se hace una extracci\'on de caracter\'isticas y una reducci\'on de dimensionalidad de los datos usando los m\'etodos de \emph{Random Forest} (RF) y correlaci\'on en conjunto, logrando reducir de $43$ a $11$ variables.
En la segunda etapa (letras de color rojo), se busca hallar las variables m\'as predictivas usando regresi\'on log\'istica, obteni\'endose $6$ variables finales. Estas variables son usadas en la \'ultima etapa (color negro),  para detectar el trastorno SLI en ni\~nos a partir de transcripciones de narrativas espont\'aneas usando k-NN. A continuaci\'on se introducen los m\'etodos del \emph{pipeline} propuesto, siguiendo la siguiente nomenclatura:

Sea $X$ la matriz que contiene los datos de tama\~no $N\times V$, donde $N$ es cantidad de observaciones y $V$ la cantidad de variables. Note que $x$ corresponde a una observaci\'on de una variables  espec\'ifica. Recordar que el objetivo final, es la detecci\'on del trastorno espec\'ifico del lenguaje (SLI)  partir de transcripciones de narrativas espont\'aneas, este objetivo se enmarca en un problema de clasificaci\'on binaria, por ende es necesario considerar dos clases $C=0$ para un desarrollo t\'ipico normal y $C=1$  para SLI.

\subsection{Lenguaje de programaci\'on}
\label{ssec:soft}
La implementaci\'on de los siguientes m\'etodos fueron realizados usando el lenguaje de programaci\'on  \emph{RStudio 2023.09.1+494, Desert Sunflower Release}, el cual est\'a desarrollado para computaci\'on estad\'istica y visualizaci\'on de datos.

\subsection{Base de Datos}
\label{ssec:db}
La base de datos consiste en transcripciones de audio p\'ublicas de tres estudios diferentes, llamados: Conti-Ramsden $4$, ENNI y GILLUM. Se puede accesar libremente en \cite{database}.

El conjunto de datos Conti-Ramsden $4$ se recopil\'o para un estudio que evalu\'o la efectividad de las pruebas narrativas en adolescentes. Consiste en 99 muestras de desarrollo t\'ipico (TD) y 19 muestras de trastorno espec\'ifico del lenguaje (SLI) de ni\~nos entre las edades de $13$ y $16$. Este contiene transcripciones de una tarea de narraci\'on basada en el libro de im\'agenes sin palabras. 
El conjunto de datos ENNI  consta de $300$ muestras de desarrollo t\'ipico (TD) y $77$ muestras de trastorno espec\'ifico del lenguaje (SLI) de ni\~nos entre $4$ y $9$ a\~nos. A cada ni\~no se le presentaron dos historias de im\'agenes sin palabras, una m\'as complicada que la otra. 
El conjunto de datos de Gillam se basa en otra herramienta para la evaluaci\'on narrativa conocida como \emph{ Test de Lenguaje Narrativo (TNL)}. Consiste en $250$ ni\~nos con trastornos del lenguaje (SLI) y $520$ TD de entre $5$ y $12$ a\~nos. 

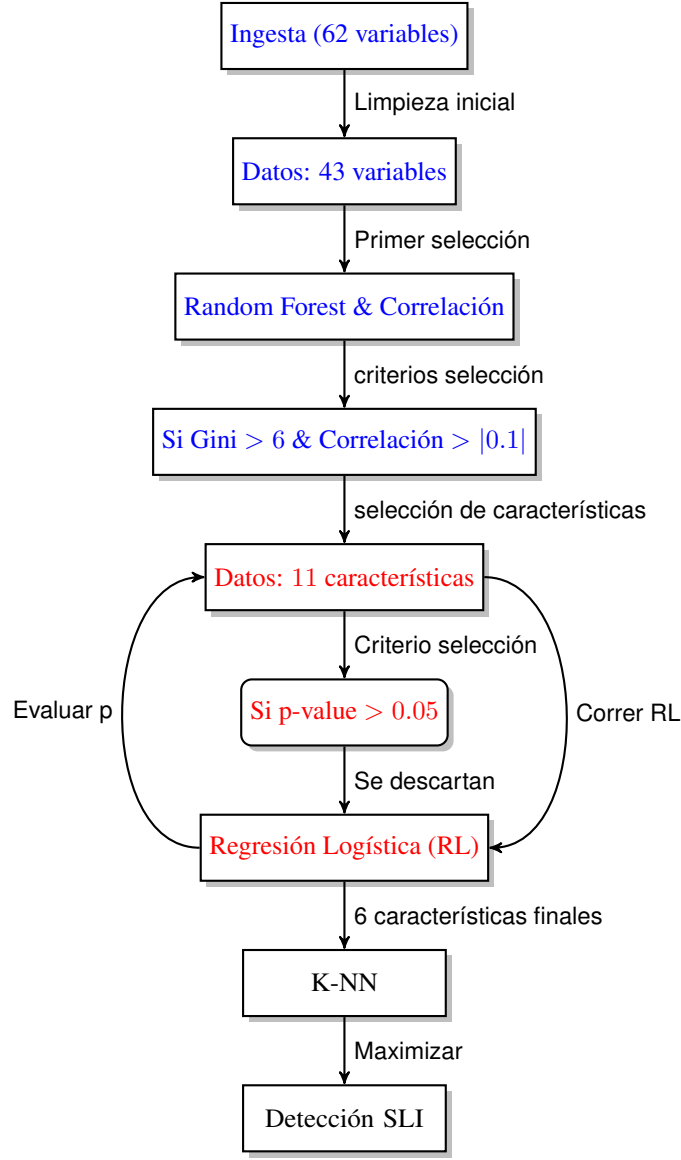
\begin{figure}[hbt]
		\centering
		\tikzstyle{none}   = [above, text centered]
		\tikzstyle{block}  = [draw, fill=white, text centered, minimum height=2.5em, minimum width=2.5em,drop shadow]
		\tikzstyle{block1} = [block, fill=white!30, minimum width=3em]
		\tikzstyle{block2} = [block, fill=white!30,  minimum width=3em, minimum height=2.5em, rounded corners]
		\begin{tikzpicture}[
			->,>=stealth',auto,thick, node distance=1.8cm]
			\node[block1] (dato) {\hlb{Ingesta (62 variables)}};
			\node[block1] (clean) [below of=dato] {\hlb{Datos: 43 variables}};
			\node[block1] (rf)    [below of=clean] {\hlb{Random Forest \& Correlaci\'on}};
                \node[block1] (gini)  [below of=rf] {\hlb{Si Gini $> 6$ \& Correlaci\'on $>|0.1|$}};
                \node[block1] (corr)  [below of=gini] {\hlr{Datos: $11$ caracter\'isticas}};   
                \node[block2] (pvalue)[below of=corr] {\hlr{Si p-value $>0.05$}};
                \node[block1] (reglog)[below of=pvalue] {\hlr{Regresi\'on Log\'istica (RL)}};
                \node[block1] (knn)   [below of=reglog,text width=7em] {K-NN};
                \node[block1] (sli)   [below of=knn,text width=7em] {Detecci\'on SLI};
			
  			\path[every node/.style={font=\sffamily\small}]
				(dato)   edge node [right] {Limpieza inicial} (clean) 
				(clean)  edge node [right] {Primer selecci\'on} (rf)
				(rf)     edge node [right] {criterios selecci\'on} (gini)
                    (gini)   edge node [right] {selecci\'on de caracter\'isticas} (corr)
                    (corr)   edge node [right] {Criterio selecci\'on} (pvalue)
                    (pvalue) edge node [right] {Se descartan} (reglog)
                    (reglog) edge node [right] {6 caracter\'isticas finales} (knn)
                    (knn)    edge node [right] {Maximizar} (sli)

                (corr)   edge[out=0, in=0] node [right] {Correr RL} (reglog)
   			(reglog) edge[out=180, in=180] node [left] {Evaluar p} (corr)
               ;
		\end{tikzpicture}
\caption{\emph{Pipeline} que ilustra la metodolog\'ia propuesta. }
\label{fig:bloques}
\end{figure}
Las bases de datos combinadas contiene $1163$ observaciones y $62$ variables. 
Dentro de estas variables se encuentra el diagn\'ostico correspondiente a cada ni\~no, si este tiene un desarrollo t\'ipico o muestra SLI, esta variable corresponde a la variable objetivo.
A modo de simplificaci\'on se decidi\'o retirar aquellas variables que no fueron una m\'etrica num\'erica medible, c\'omo puede ser un conteo de palabras o s\'ilabas. De esta manera se cuentan con variables que registran el desempe\~no de un ni\~no descomponiendo su narrativa en la calidad de sus enunciados, oraciones y palabras. Algunos ejemplos son la cantidad de palabras de relleno por oraci\'on, la cantidad de errores por palabra, el promedio de verbos en pasado sobre los verbos en infinitivo, el promedio de palabras,verbos o adjetivos por oraci\'on,  entre otros. El objeto de estas observaciones consiste en evaluar la complejidad narrativa y en ella, poder predecir un desarrollo at\'ipico en la misma. Escapa el objeto de la presente investigaci\'on indagar en profundidad sobre los patrones del lenguaje en ni\~nos y sus variantes; no obstante, se invita al lector a explorar los siguientes autores en la materia de an\'alisis de narrativas y extracci\'on de caracter\'isticas \cite{brown1973first,bowen1998}.

\subsection{Extracci\'on de caracter\'isticas}
\label{ssec:car}
Se entiende por Extracci\'on o selecci\'on de caracter\'isticas a la pr\'actica de reducir a un conjunto de caracter\'isticas con el objeto de mejorar una inducci\'on \cite{wu2009top}.
Estos m\'etodos se utilizan para obtener el subconjunto de caracter\'isticas m\'as relevantes del conjunto principal, dicho subconjunto ser\'a aquel que maximice una funci\'on criterio
determinada \cite{violini2014}.
En esta etapa se consider\'o la relaci\'on entre la habilidad narrativa y el lenguaje. Espec\'ificamente se considera como un m\'etodo v\'alido la medici\'on de la competencia comunicativa de un individuo \cite{Botting2008}. Para lograr esto, solo se consideraron las variables cuantitativas relacionadas con la narrativa de los ni\~nos, dejando fuera aquellas que no representan un dato concreto y medible. Esta selecci\'on fue hecha manualmente obteni\'endose $43$ variables iniciales. 
El siguiente paso fue aplicar un esquema para determinar la importancia de estas variables junto con su capacidad de predicci\'on de la variable objetivo, dada por la variable grupo (ver Secci\'on \ref{ssec:db}). Para dicho fin se usaron los m\'etodos de \emph{Random Forest} y Correlaci\'on.

\vspace*{1em}
\underline{Random Forest} (RF): Es un modelo utilizado como un m\'etodo de clasificaci\'on y regresi\'on de prop\'osito general. El enfoque, que combina varios \'arboles de decisi\'on aleatorizados y agrega sus predicciones promediando, es lo suficientemente vers\'atil como para ser aplicado a problemas a gran escala, se adapta f\'acilmente a diversas tareas de aprendizaje \emph{ad-hoc} y proporciona medidas de importancia de variables \cite{Biau2016}. Como clasificador binario, se centra en una votaci\'on mayoritaria entre los \'arboles de clasificaci\'on \cite{Quintero2019}. Esto significa que para una observaci\'on $x$, se predice $Y = 1$ si m\'as de la mitad de los \'arboles individuales predicen $Y = 1$, y $Y = 0$ en caso contrario. Esto se puede expresar de la siguiente manera:
\begin{align}
    \label{eq:rf}
    m_M,n(x; \theta_1, \cdots, \theta_M, &D_n) = 
    \begin{cases} 
    1 & \text{si } \frac{1}{M} \sum_{j=1}^{M} m_n(x; \theta_j, D_n) > \frac{1}{2} \\ 0 & \text{en otro caso} 
    \end{cases}
\end{align}

La funci\'on \eqref{eq:rf} calcula el voto mayoritario de $M$ modelos, cada uno con sus par\'ametros  $\theta_1, \cdots, \theta_M$, entrenados con el conjunto de datos $D_n$. La sumatoria calcula la media de las predicciones de cada modelo $m_n(x; \theta_j, D_n)$ para la entrada $x$ sobre el conjunto de modelos $M$. La funci\'on de decisi\'on determina la clase final asignada a la instancia $x$, basada en si la media de las predicciones supera o no, el valor de $0.5$ \cite{Breiman2001}.  El n\'umero de \'arboles  $M$ puede elegirse arbitrariamente grande, entonces, desde un punto de vista de modelado, en \eqref{eq:rf} $M$ puede tender a $\infty$, por lo tanto:

\begin{align}
\label{eq:rf2}
    m_{\infty,n}(x; D_n) = E_{\theta} [m_n(x; \theta, D_n)] .
\end{align}

En esta definici\'on, $E_{\theta}$ denota la esperanza con respecto al par\'ametro aleatorio  $\theta$, condicionado a $D_n$, entonces \eqref{eq:rf2}, se puede expresar como:
\begin{align}
\label{eq:rf3}
    \lim_{M \rightarrow \infty} m_{M,n}(x; \theta_1, \ldots, \theta_M, D_n) = m_{\infty,n}(x; D_n)
\end{align}

Luego se usa el criterio de \'arboles de clasificaci\'on y regresi\'on (\emph{CART}, por sus siglas en ingl\'es \emph{Classification and Regression Trees}) para realizar el ajuste binario. B\'asicamente, se define como una funci\'on de la impureza de Gini en cada nodo del \'arbol de decisi\'on. Matem\'aticamente esta impureza se puede escribir como:
\begin{align}
\label{eq:impgini}
    Gini = 1 -\sum_{i=1}^{n} (p_i)^{{2}}    
\end{align}
donde $p_i$ es la probabilidad de que un objeto sea clasificado en una clase particular. Esta medida tiene en cuenta la proporci\'on de clases en cada nodo y se utiliza para determinar la mejor manera de dividir el conjunto de datos en subconjuntos m\'as puros. Se puede expresar como:
\begin{align}
    \label{eq:gini}
    L_{\text{class},n}(j, z) &= p_{0,n}(A)p_{1,n}(A) - \frac{N_{n}(AL)}{N_{n}(A)}p_{0,n}(AL)p_{1,n}(AL) 
    \nonumber
    \\&
    - \frac{N_{n}(AR)}{N_{n}(A)}p_{0,n}(AR)p_{1,n}(AR)
\end{align}

En \eqref{eq:gini}, el nodo $n$ consiste en un \'arbol de decisi\'on, al considerar una divisi\'on $j$ con el umbral $z$. En ella:
\begin{itemize}
    \item $p_{0,n}(A)$ y $p_{1,n}(A)$ denotan las proporciones de las clases $0$ y $1$ respectivamente en el nodo $n$.
    \item $N_{n}(AL)$ y $N_{n}(AR)$ representan el n\'umero de instancias en las ramas izquierda ($AL$) y derecha ($AR$) del nodo $n$, despu\'es de aplicar la divisi\'on $j$ con el umbral $z$. 
    \item $p_{0,n}(AL)$ y $p_{1,n}(AL)$,  indican las proporciones de las clases $0$ y $1$ en la rama izquierda del nodo $n$ despu\'es de la divisi\'on.
    \item $p_{0,n}(AR)$ y $p_{1,n}(AR)$ representan las proporciones de las clases $0$ y $1$ en la rama derecha del nodo $n$ despu\'es de la divisi\'on. 
\end{itemize}
La ecuaci\'on \eqref{eq:gini} se emplea para calcular la ganancia de informaci\'on o la reducci\'on de la impureza al dividir un nodo en dos ramas \cite{Biau2016}.

\vspace*{1em}
\underline{Coeficiente de correlaci\'on de \emph{Spearman}} ($r_s$): Eval\'ua la relaci\'on monot\'onica entre dos variables, indicando c\'omo una variable cambia consistentemente cuando la otra aumenta o disminuye. $r_s$ est\'a basado en el coeficiente de correlaci\'on de \emph{Spearman} ($d_i$), el cual denota la diferencia entre los rangos de las observaciones $x_i$ e $y_i$, donde $n$ es el n\'umero de observaciones. Los rangos de las observaciones, $\text{rg}(x_i)$ y $\text{rg}(y_i)$, son asignados a las variables $x_i$ e $y_i$ respectivamente.

\begin{align}
\label{eq:corr}
    r_s &= 1 - \frac{6 \sum_{i=1}^{n} d_i^2}{n(n^2 - 1)}
\end{align}

Este coeficiente proporciona un valor en el rango de $-1$ a $1$, donde $-1$ indica una correlaci\'on negativa perfecta, $0$ indica ausencia de correlaci\'on, y $1$ indica una correlaci\'on positiva perfecta. El coeficiente de \emph{Spearman} es \'util dado que no asume una relaci\'on lineal entre las variables \cite{becker1988}.


\vspace*{1em}
\underline{Regresi\'on log\'istica}: Es un modelo de clasificaci\'on utilizado para predecir la probabilidad de que una variable categ\'orica dependiente tenga un valor espec\'ifico en funci\'on de una o m\'as variables independientes \cite{Cheng2006}. 
La primera expresi\'on, la funci\'on sigmoide, com\'unmente utilizada en la regresi\'on log\'istica. Toma cualquier valor de entrada $x$ y lo transforma en un valor entre $0$ y $1$. Esto es \'util para modelar probabilidades, ya que puede interpretarse como la probabilidad de que ocurra un evento binario dado un conjunto de caracter\'isticas. Esta ocurrencia se entiende como $P(y=k|x)$, la probabilidad de que la observaci\'on pertenezca a la clase $k$ dada la entrada $x$. 
\begin{align}
\label{eq:reglog}
    f(x) = \frac{1}{1 + e^{-x}} \quad \text{y} \quad P(y=k | \mathbf{x}) = \frac{e^{\mathbf{x}^T\mathbf{w}_k}}{\sum_{j=1}^{K} e^{\mathbf{x}^T\mathbf{w}_j}}
\end{align}
Donde $P(y=k|x)$ es la probabilidad de que la variable de respuesta $y$ tome el valor $k$ dado el vector de caracter\'isticas $x$, $\mathbf{w}_k$ son los par\'ametros asociados con la clase $k$, $\mathbf{x}^T$ representa la transposici\'on del vector de caracter\'isticas, y $\sum_{j=1}^{K}$ denota la suma sobre todas las clases $K$. En la sumatoria, $j$ representa cada una de las $K$ clases distintas en el problema de clasificaci\'on.  La funci\'on \emph{softmax} garantiza que la suma de todas las probabilidades sea igual a 1, lo que la hace adecuada para problemas de clasificaci\'on multiclase \cite{Zakharov2011,Flach2012ML}.

\vspace*{1em}
\underline{$p$-value o valor $p$}: Es una medida estad\'istica que indica la probabilidad de obtener resultados igualmente extremos o m\'as extremos que los observados, bajo la suposici\'on de que la hip\'otesis nula es verdadera. En el contexto de la regresi\'on log\'istica, el \textit{p-value} se utiliza para evaluar la significancia estad\'istica de cada coeficiente estimado en el modelo.

\begin{align}
\label{eq:pvalue}
    \text{\textit{$p$-value}} = 2 \times \left(1 - \text{pnorm} \left( \frac{|\text{coeficiente}|}{\text{error est\'andar}} \right)\right)
\end{align}

Donde \emph{pnorm} es la funci\'on de distribuci\'on acumulativa normal est\'andar, \emph{coeficiente} es el valor estimado del coeficiente del predictor en el modelo de regresi\'on log\'istica y \emph{error  est\'andar} es el error est\'andar asociado al coeficiente estimado \cite{james2013P-value}.

\vspace*{1em}
\underline{Mediana} ($\tilde{x}_{0.5}$): Es el valor que divide las observaciones en dos partes iguales de manera que al menos el $50\%$ de los valores sean mayores o iguales a la mediana y al menos el $50\%$ de los valores sean menores o iguales a la mediana. Se denota como $\tilde{x}_{0.5}$; luego, en t\'erminos de la funci\'on de distribuci\'on acumulada emp\'irica, se cumple la condici\'on $F(\tilde{x}_{0.5}) = 0.5$. Sean las $n$ observaciones $x_1, x_2, \ldots, x_n$ que pueden ser ordenadas como $x(1) \leq x(2) \leq \ldots \leq x(n)$. El c\'alculo de la mediana depende de si el n\'umero de observaciones $n$ es impar o par. Cuando $n$ es impar, entonces $\tilde{x}_{0.5}$ es el valor medio ordenado. Cuando $n$ es par, entonces $\tilde{x}_{0.5}$es la media aritm\'etica de los dos valores medios ordenados \cite{Introduction_to_Statistics}.

\begin{align}
\label{eq:mediana}
\tilde{x}_{0.5} = \begin{cases} 
x\left(\frac{n+1}{2}\right) & \text{si } n \text{ es impar} \\
\frac{1}{2} \left( x\left(\frac{n}{2}\right) + x\left(\frac{n}{2}+1\right) \right) & \text{si } n \text{ es par} \\
\end{cases}
\end{align}

\vspace*{1em}
\underline{Cuartiles}: Son valores que dividen los datos en cuatro partes iguales. El primer cuartil, denotado como $Q_1$, es el valor que deja el $25\%$ de los datos a su izquierda. El segundo cuartil es la mediana $\tilde{x}_{0.5}$. El tercer cuartil, denotado como $Q_3$, es el valor que deja el $75\%$ de los datos a su izquierda. Para calcular el $p$-\'esimo cuartil, denotado como $Q_p$, en una muestra ordenada de $n$ observaciones es:
\begin{align}
    \label{eq:cuartil}
    Q_p = x\left(\frac{np}{100}\right)
\end{align}

\subsection{k-vecinos m\'as cercanos (k-NN)}
\label{ssec:knn}  
$K$-NN es un algoritmo de aprendizaje supervisado utilizado para la clasificaci\'on y regresi\'on. En la clasificaci\'on $K$-NN, se asigna una etiqueta de clase al punto de datos desconocido, bas\'andose en la mayor\'ia de las etiquetas de clase de los $k$ puntos de datos m\'as cercanos en el conjunto de entrenamiento \cite{wu2009top,Quintero2020}. Usando el vector de caracter\'isticas $\Theta_{Ct}$, consideramos una clasificaci\'on en dos posibles clases $c=0$ (Desarrollo t\'ipico) y $c=1$ (SLI). La probabilidad de clasificar una muestra en una de las dos clases est\'a dada por:

\begin{align}
    \nonumber
    \rho(\Theta_{Ct}|c=0) &= \frac{1}{N_0}\sum_{n \in \text{class } 0} \mathcal{N}(\Theta_{Ct}|\Theta_{nCt},\sigma^2I) \\
    &= \frac{1}{N_0(2\pi\sigma^2)^{D/2}}\sum_{n \in \text{class } 0} \exp\left(-\frac{(\Theta_{Ct}-\Theta_{nCt})^2}{2\sigma^2}\right) \\
    \nonumber
    \rho(\Theta_{Ct}|c=1) &= \frac{1}{N_1}\sum_{n \in \text{class } 1} \mathcal{N}(\Theta_{Ct}|\Theta_{nCt},\sigma^2I) \\
    &= \frac{1}{N_1(2\pi\sigma^2)^{D/2}}\sum_{n \in \text{class } 1} \exp\left(-\frac{(\Theta_{Ct}-\Theta_{nCt})^2}{2\sigma^2}\right)
\end{align}

donde $D$ es la dimensi\'on de la muestra $\Theta_{Ct}$, $N_0$ y $N_1$ son los n\'umeros de muestras de entrenamiento de la clase $0$ y clase $1$, respectivamente, y $\sigma^2$ es la varianza. Usando la regla de Bayes para clasificar una nueva observaci\'on $\Theta^*_{Ct}$, obtenemos la siguiente ecuaci\'on:

\begin{align}
\label{eq:knn1}
    \rho(c=0|\Theta^*_{Ct}) &= \frac{\rho(\Theta^*_{Ct}|c=0)\rho(c=0)}{\rho(\Theta^*_{Ct}|c=0)\rho(c=0) + \rho(\Theta^*_{Ct}|c=1)\rho(c=1)},
\end{align}

donde la m\'axima verosimilitud nos da $\rho(c=0) = N_0/(N_0 + N_1)$ y $\rho(c=1) = N_1/(N_0 + N_1)$. Sustituyendo en la ecuaci\'on \eqref{eq:knn1}, obtenemos la probabilidad $\rho(c=0|\Theta_{Ct})$. La expresi\'on para $\rho(c=1|\Theta_{Ct})$ se puede derivar de manera similar. Para determinar qu\'e clase es m\'as probable, se eval\'ua la proporci\'on entre las dos expresiones:

\begin{align}
\label{eq:knn2}
\frac{\rho(c=0|\Theta_{Ct})}{\rho(c=1|\Theta_{Ct})} &= \frac{\rho(\Theta^*_{Ct}|c=0)\rho(c=0)}{\rho(\Theta^*_{Ct}|c=1)\rho(c=1)},
\end{align}

Si la proporci\'on es mayor que uno, $\Theta_{Ct}$ se clasifica como $c=0$, de lo contrario se clasifica como $c=1$. Es importante se\~nalar que en el caso donde $\sigma^2$ es muy peque\~no en (11), entonces tanto el numerador como el denominador estar\'an dominados por el t\'ermino para el cual la muestra $\Theta_{n0Ct}$ en la clase-0 o $\Theta_{n1Ct}$ en la clase-1 est\'an m\'as cerca del punto $\Theta_{Ct}$, tal que:

\begin{align}
    \nonumber
    \frac{\rho(c=0|\Theta_{Ct})}{\rho(c=1|\Theta_{Ct})} &= \frac{\exp\left(-\frac{(\Theta^*_{Ct}-\Theta_{n0Ct})^2}{2\sigma^2}\right) \rho(c=0)/N_0}{\exp\left(-\frac{(\Theta_{Ct}-\Theta_{n1Ct})^2}{2\sigma^2}\right) \rho(c=1)/N_1} \\
&= \frac{\exp\left(-\frac{(\Theta_{Ct}-\Theta_{n0Ct})^2}{2\sigma^2}\right)}{\exp\left(-\frac{(\Theta^*_{Ct}-\Theta_{n1Ct})^2}{2\sigma^2}\right)}.
\end{align}

En el l\'imite $\sigma^2 \rightarrow 0$, $\Theta_{Ct}$ se clasifica como clase 0 si $\Theta_{Ct}$ tiene un punto en los datos de clase $0$ que est\'a m\'as cerca que el punto m\'as cercano en los datos de clase $1$. El m\'etodo del vecino m\'as cercano se recupera as\'i como el caso l\'imite de un modelo generativo probabil\'istico. El par\'ametro $k$ se elige basado en $N^{1/2}$, donde $N$ es el n\'umero de muestras en el conjunto de datos de entrenamiento. Para un tratamiento completo de las propiedades matem\'aticas del clasificador de $k$-vecinos m\'as cercanos, remitimos al lector a \cite{bishop2006,barber2012bayesian}.

\subsection{Validaci\'on cruzada qu\'intuple}
\label{ssec:valcr}
La validaci\'on cruzada es un m\'etodo utilizado para evaluar el rendimiento predictivo de un modelo de aprendizaje autom\'atico \cite{Kohavi1995}. Consiste en dividir el conjunto de datos en subconjuntos de entrenamiento y prueba repetidamente, ajustando y evaluando el modelo en cada iteraci\'on. Esta dada por:

\begin{align}
\label{eq:vc5}
    \bar{p} = \frac{1}{k} \sum_{i=1}^{k} p_{i}    
\end{align}
donde $k$ es el n\'umero de iteraciones de la validaci\'on cruzada, $\bar{p}$ es la precisi\'on media, $p_{i}$ es la precisi\'on del modelo en la $i$-\'esima iteraci\'on.

\subsection{M\'etricas de evaluaci\'on}   
\label{ssec:metricas}
Se refieren a medidas utilizadas para evaluar el rendimiento de los modelos de aprendizaje autom\'atico. Estas m\'etricas permiten cuantificar la calidad de las predicciones realizadas por un modelo en funci\'on de los datos de prueba \cite{Flach2012ML}. A continuaci\'on se introducen las m\'etricas usadas en este estudio.
 
\underline{Precisi\'on}: Es la proporci\'on de instancias positivas clasificadas correctamente entre todas las instancias clasificadas como positivas por el modelo. Se calcula mediante la f\'ormula:
\begin{align}
\label{eq:pres}
    \text{Precision} = \frac{TP}{TP + FP}
    \end{align}
donde $TP$ es n\'umero de verdaderos positivos y FP es n\'umero de falsos positivos. Se entiende por TP a los casos en los que el modelo clasifica correctamente una instancia como positiva cuando realmente lo es. Por otra parte, $FP$ son los casos en los que el modelo clasifica incorrectamente una instancia como positiva cuando en realidad es negativa.

\underline{Recall positivo}: Sensibilidad, es la proporci\'on de instancias positivas clasificadas correctamente entre todas las instancias que realmente son positivas. Se calcula mediante la f\'ormula:
\begin{align}
\label{eq:recall}
\text{Recall} = \frac{TP}{TP + FN}
\end{align}
con $FN$ siendo n\'umero de falsos negativos, donde modelo clasifica incorrectamente una instancia como negativa cuando en realidad es positiva. Por otra parte, los Falso Negativo (FN) son aquellos casos donde el modelo clasifica incorrectamente una instancia como negativa cuando en realidad es positiva. 

\underline{Recall negativo}: Especificidad, se refiere a la proporci\'on de instancias negativas clasificadas correctamente entre todas las instancias que realmente son negativas. Se entiende como . Se calcula mediante la f\'ormula:
\begin{align}
\label{eq:nrecall}
\text{Neg-Recall} = \frac{TN}{TN + FP}
\end{align}

\underline{F1-score}: Es la media arm\'onica de precisi\'on y recall y proporciona un equilibrio entre ambas m\'etricas. Se calcula mediante la f\'ormula:
\begin{align}
\label{eq:f1}
F1 = 2 \times \frac{\text{Precision} \times \text{Recall}}{\text{Precision} + \text{Recall}}
\end{align}

\underline{\'area bajo la Curva, $ROC$}: (AUC-ROC) es una medida de la capacidad discriminatoria de un modelo de clasificaci\'on binaria. Representa la probabilidad de que el modelo clasifique correctamente una instancia positiva al azar m\'as alta que una instancia negativa al azar.
\begin{align}
    AUC = \int_{0}^{1} \text{Sensibilidad} \cdot d(\text{Especificidad})    
\end{align}
La $ROC$, tiene una relaci\'on directa con la sensibilidad y la especificidad, estas se calculan a partir de la matriz de confusi\'on obtenida al evaluar el modelo con datos de prueba. \cite{Krupinski2017,Hanley1982Roc}

\underline{Error de ra\'iz cuadr\'atica media} ($RMSE$): Es una medida que cuantifica la diferencia entre los valores predichos por un modelo y los valores observados. Se calcula tomando la ra\'iz cuadrada de la media de los cuadrados de las diferencias entre los valores observados y los valores predichos por el modelo. Esta m\'etrica se utiliza para evaluar la precisi\'on del modelo en t\'erminos de unidades de la variable dependiente. 
\begin{align}
    \label{eq:rmse}
    RMSE = \sqrt{\frac{1}{n} \sum_{i=1}^{n}(y_i - \hat{y}_i)^2}
\end{align}

\underline{Error medio absoluto} ($MAE$): Es una medida que calcula la magnitud promedio de los errores en las predicciones de un modelo, sin tener en cuenta su direcci\'on. Se obtiene calculando la media de las diferencias absolutas entre los valores observados y los valores predichos por el modelo. Esta m\'etrica proporciona una medida de la magnitud promedio de los errores de predicci\'on \cite{Flach2012ML}.
\begin{align}
\label{eq:mae}
    MAE = \frac{1}{n} \sum_{i=1}^{n} |y_i - \hat{y}_i|
\end{align}

\underline{$R^2$}: Es una medida que indica la proporci\'on de la varianza en la variable dependiente que es predecible a partir de la(s) variable(s) independiente(s) en el modelo. Se calcula como $1$ menos la proporci\'on de la varianza residual respecto a la varianza total. Un $R^2$ m\'as alto indica un mejor ajuste del modelo a los datos observados \cite{Flach2012ML}.
\begin{align}
\label{eq:r2}
    R^2 = 1 - \frac{\sum_{i=1}^{n}(y_i - \hat{y}_i)^2}{\sum_{i=1}^{n}(y_i - \bar{y})^2}
\end{align}
Donde $n$ es el n\'umero de observaciones, $y_i$ es el valor observado, $\hat{y}_i$ es el valor predicho por el modelo para la $i$-\'esima observaci\'on, e $\bar{y}$ es la media de los valores observados \cite{Flach2012ML}.

\underline{Error $OOB$}: Se calcula utilizando la tasa de error (clasificaci\'on) o el error cuadr\'atico medio (regresi\'on) de las predicciones hechas en las muestras OOB (por sus siglas en ingl\'es \emph{Out of the Bag Error}). Para un problema de clasificaci\'on binaria, la f\'ormula del error OOB se puede expresar como:
\begin{align}
\label{eq:oob}
\text{Error OOB} = \frac{1}{N} \sum_{i=1}^{N} I(y_i \neq \hat{y}_i)
\end{align}
Donde $N$ es el n\'umero de muestras en el conjunto de datos, $y_i$ es la etiqueta verdadera de la muestra $i$, $\hat{y}_i$ es la predicci\'on del modelo para la muestra $i$, $I$ es una funci\'on indicadora que devuelve $1$ si la condici\'on dentro de los par\'entesis es verdadera, y $0$ en caso contrario \cite{james2013P-value}.

\section{Resultados y discusi\'on}
\label{sec:res}
En la presente secci\'on se discuten los resultados hallados tras aplicar el \emph{pipeline} propuesto siguiendo la Figura~\ref{fig:bloques} de la Metodolog\'ia, Secci\'on \ref{sec:met}. Todo sobre las $1063$ observaciones y las $43$ variables relacionadas con las narrativas espont\'aneas en ni\~nos, introducidas en la Secci\'on \ref{ssec:db}.

\begin{figure}[H]
    \centering
    \includegraphics[angle=0,width=0.7\textwidth]{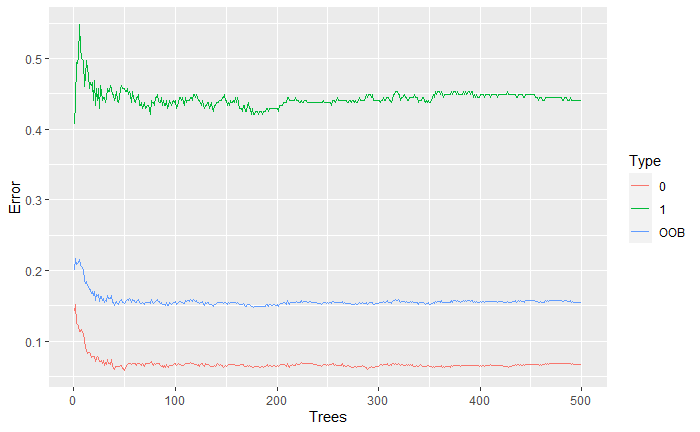}
    \caption{Comportamiento del modelo con respecto a cada clase de predicci\'on. Observe como los \'arboles se empiezan a estabilizar a partir del valor $16$ aproximadamente, para el tipo $1$ (Clase $1$) y $OOB$, mientras que para el tipo $0$ (Clase $0$), se empieza a estabilizar en $200$ aproximadamente. Esta inspecci\'on visual permite establecer el valor de los \'arboles en $500$.}
    \label{fig:arb}
\end{figure}

La primera etapa consiste en reducir las cantidad de datos. Entonces, como primer instancia se necesitan estimar los par\'ametros del m\'etodo RF, dado por la ecuaci\'on \eqref{eq:rf}, para as\'i determinar, cu\'ales son las variables m\'as importantes de los datos. Recordar que RF como clasificador binario, se centra en la votaci\'on mayoritaria de los \'arboles, por lo tanto, es necesario estimar la cantidad \'arboles \'optimos del m\'etodo. Para dicho fin, se busca el valor \'optimo OOB, ver ecuaci\'on \eqref{eq:oob}, entre las $43$ variables resultantes de limpiar los datos originales. La idea es encontrar el par\'ametro relacionado con el n\'umero de variables aleatorias, como candidatas en cada ramificaci\'on que registre el OOB \'optimo.  La Figura~\ref{fig:arb} muestra como var\'ia el error OOB con el n\'umero de \'arboles y muestra c\'omo se comporta el modelo con respecto a cada clase de predicci\'on. Observe los valores de los \'arboles son bastante constantes desde el valor $50$, para el tipo $1$ y para OOB, m\'as sin embargo, se estabilizan en el valor $500$. Teniendo el valor de la cantidad de \'arboles $M$, es posible entonces, estimar el m\'etodo RF para las clases $C=0=normal$ y $C=1=SLI$ usando la ecuaci\'on \eqref{eq:rf3}. 

\begin{figure}[H]
\centering
\subfigure[Boxplot] {\includegraphics[angle=0,width=75mm]{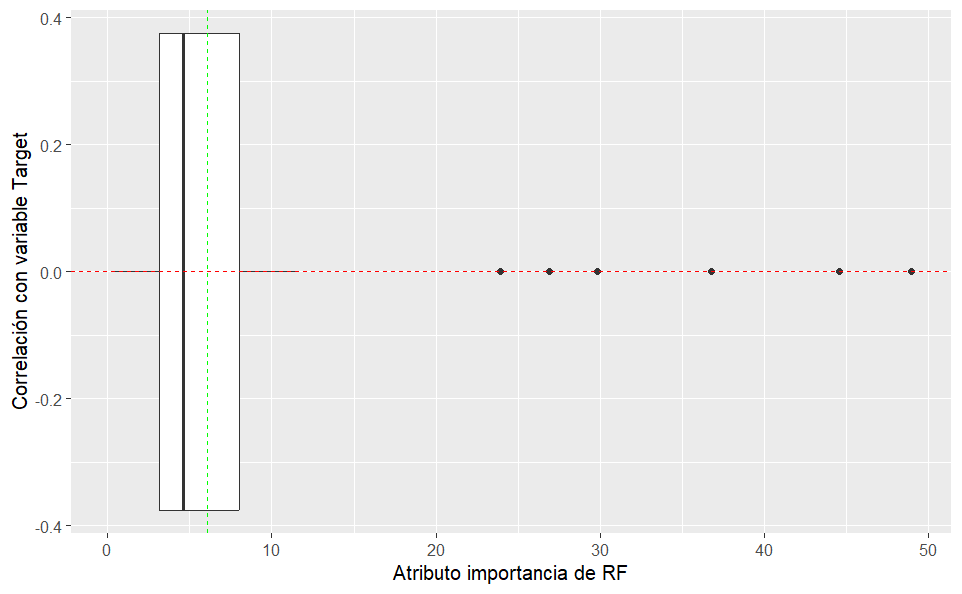} \label{fig:rfbox}}
\subfigure[Gr\'afico de dispersi\'on]{\includegraphics[angle=0,width=75mm]{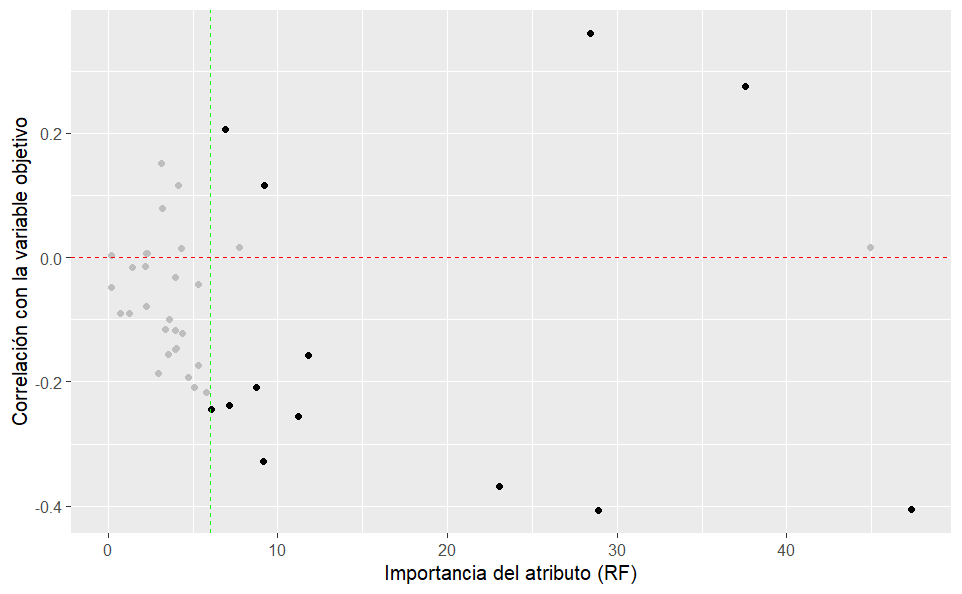}\label{fig:rfycor}}
\caption{Comportamiento para el atributo de importancia de RF y correlaci\'on usando (a)~{Boxplot} y (b)~{Gr\'afico de dispersi\'on}. Se puede observar los criterios de selecci\'on en rojo y verde en ambos gr\'aficos. Siendo el de correlaci\'on nula en rojo y de atributo importancia de \textit{Random Forest} mayor a seis, en verde. Los puntos grises consisten variables descartados, en tanto los negros, caracter\'isticas }
\label{fig:rfcorr}
\end{figure}

La meta en esta etapa, ~es encontrar la selecci\'on de caracter\'isticas de mayor importancia y as\'i hacer una primera reducci\'on de dimensionaliad de los datos. Esta parte involucra el m\'etodo RF con el \emph{atributo importancia}, junto con el m\'etodo del coeficiente de correlaci\'on con la variable \emph{objetivo}. Este criterio de selecci\'on se compone de dos partes. 
\begin{enumerate}
    \item Criterio 1:  Consiste en elegir aquellas variables que cumplan tener, una importancia superior al promedio aproximado entre la mediana y el tercer cuartil, cuyos valores son $4,5$ y $7,8$ respectivamente, ver ecuaciones \eqref{eq:mediana} y \eqref{eq:cuartil}. La raz\'on de esta decisi\'on consiste en aprovechar no solo el mejor $25\%$ de los datos (por delante del tercer cuartil), si no tambi\'en, aquellas variables que se encontraban entre la mediana y el punto de corte del mencionado cuartil. Este enfoque permite aportar caracter\'isticas de valor, que de otra forma hubiesen sido descartadas. En la Figura~\ref{fig:rfbox} es posible apreciar una importante aglomeraci\'on antes del punto de corte (linea rojaerde) entre la mediana y el tercer cuartil. Donde el eje X es el atributo importancia de RF y en el eje y, la correlaci\'on con la variable \textit{objetivo}. Por otra parte en la figura \ref{fig:rfycor} se aprecia la utilidad de la propuesta planteada, en tanto se identifican aquellas variables que no poseen una importancia significativa para el random forest (eje x). De esta manera, se conservan variables de importancia sin arrastrar consigo aquellas de menor atributo de importancia, las cuales se encuentran alrededor de la mediana, por la izquierda de la linea verde. 
    \item Criterio 2: Consiste en corroborar que aquellas variables cuyo coeficiente de Gini, dado por la ecuaci\'on \eqref{eq:impgini}, sea mayor a $6$. este valor se obtiene al aplicar la ecuaci\'on \eqref{eq:gini}, y que tengan una correlaci\'on no-nula con la variable \emph{objetivo}, es decir, una $r_s>|0.1|$, ver ecuaci\'on \eqref{eq:corr}. 
    Este enfoque detect\'o $13$ variables como las m\'as importantes, m\'as, sin embargo, al analizar las correlaciones, se observ\'o que todas las variables contaban con correlaciones v\'alidas distintas de cero salvo dos, identificadas como \emph{Rellenos} y \emph{Edad}, cuyas correlaciones eran de $0.015$ y $0.016$ respectivamente. De esta manera se separan $11$ variables que pasan a ser las caracter\'isticas finales.  En La Figura~\ref{fig:rfycor} se puede observar la \emph{importancia del atributo RF} contra la \emph{correlaci\'on de la variable objetivo}. Note como se crean dos l\'ineas de intersecci\'on que hacen que se puedan separar las variables m\'as importantes (puntos de color negro). La l\'inea de corte en la correlaci\'on (l\'inea de color rojo) y el atributo importancia (l\'inea de color verde) permiten seleccionarlas caracter\'isticas m\'as importantes. Finalmente, se obtienen las siguientes 11 caracter\'isticas:  Verbos sin declinar, Media de los morfemas por oraci\'on, Errores de palabras, Promedio de s\'ilabas por palabra, Frecuencia de tipos de palabras, Uso regular del pasado, Media de las palabras por oraci\'on, N\'umero de etiquetas y n\'umero de palabras.
\end{enumerate}

La siguiente etapa del \emph{pipeline} consiste en evaluar la predictibilidad de las $11$ caracter\'isticas seleccionadas en la etapa anterior, aplicando el modelo de regresi\'on log\'istica de manera consecutiva, ver ecuaci\'on \eqref{eq:reglog}. Este enfoque, permite ir descartando aquellas variables que no consiguieran un $p$-value menor a $0.05$, ver ecuaci\'on \eqref{eq:pvalue}. El experimento se repite c\'iclicamente, hasta ya no poder conseguir descartar m\'as caracter\'isticas.  Recordar que la variable objetivo en este caso es la que esta etiquetada ya sea como un TD o SLI. Este proceso permiti\'o seleccionar un conjunto final de $6$ caracter\'isticas relevantes para el an\'alisis. En la Tabla~\ref{tab:Reglog} se detallan estas caracter\'isticas: Relaci\'on de uso de verbos sin declinar ante los declinados, Longitud media de morfemas por oraci\'on, Errores de palabras identificados en las transcripciones, Promedio de s\'ilabas por palabra, Frecuencia de tipos de palabras respecto al n\'umero total de palabras y uso regular del pasado. Adem\'as, en la tabla se presentan los estimados, errores est\'andar y valores $z$ asociados a cada una de estas caracter\'isticas, que son medidas fundamentales para comprender su contribuci\'on al modelo predictivo y su significancia estad\'istica en la clasificaci\'on de ni\~nos con trastornos espec\'ificos del lenguaje (SLI).

\begin{figure}[H]
    \centering
    \includegraphics[width=0.7\textwidth]{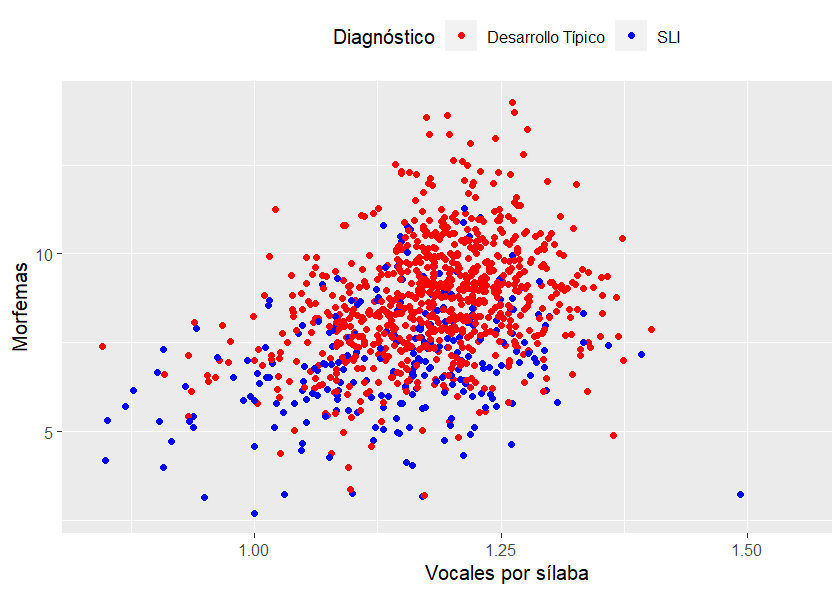} 
    \caption{Relaci\'on entre Cantidad de \emph{Morfemas} por \emph{Respuesta y promedio de silabas}.}
    \label{fig:MorfySyl}
\end{figure}

Finalmente, consid\'erese dos posibles clases $c=0$ (Desarrollo t\'ipico, TD) y $c=1$ (trastorno espec\'ifico del lenguaje, SLI) para el vector de caracter\'isticas $\Theta_{Ct}$ dado por las $6$ caracter\'isticas dadas por al etapa anterior. Se propone usar $14$-NN introducido en la Secci\'on \eqref{ssec:knn} como clasificador para detectar SLI a partir de transcripciones de narrativas espont\'aneas. A modo de ilustraci\'on, en la Figura~\ref{fig:MorfySyl}, se puede apreciar una tendencia de los pacientes con SLI (Azul) con respecto a los pacientes con TD (Rojo), espec\'ificamente la relaci\'on entre las caracter\'isticas \emph{Vocales por s\'ilabas} vs \emph{Morfemas}. Se puede observar como la dispersi\'on de los grupos est\'an bastante superpuestos, lo que hace que el modelo 14-NN sea una buena opci\'on de clasificaci\'on. El \emph{K} seleccionado surge de aplicar la ra\'iz cuadrada al numero de observaciones que componen el conjunto de entrenamiento, el cual es el 70$\%$ de las 1063 observaciones que forman los datos \cite{bishop2006}.

\begin{table}[hb]
\caption{Resultados de aplicar regresi\'on log\'istica. El ciclo de Regresiones termina al encontrar un conjunto de caracter\'isticas que cumpla la condici\'on de $p$-value $< 0.05$. Se debe lograr ya no poder descartar elementos.}
    \centering
    \label{tab:Reglog}
    \begin{tabular}{|c|c|c|c|c|}
        \hline\hline
        Caracter\'isticas & Estimados & Std. error & $z$ value & $p>|\text{z}|$ \\
        \hline\hline
        Verbos sin declinar & 1.25727 & 0.28994 & 4.336 & $\approx 0$ \\
        \hline
        Morfemas por oraci\'on & -0.84605 & 0.08618 & -9.817 &  $\approx 0$ \\
        \hline
        Errores & 0.85750 & 0.11493 & 7.461 & $\approx 0$ \\
        \hline
        Promedio silabas & -2.64640 & 1.10947 & -2.385 & 0.01707 \\
        \hline
        Frecuencia de tipos & -2.73838 & 0.91128 & -3.005 & 0.00266 \\
        \hline
        Pasado regular & -0.05929 & 0.01623 & -3.652 & 0.00026 \\
         \hline\hline
    \end{tabular}
\end{table}

Los hiperpar\'ametros de $K$-NN se eligieron usando validaci\'on cruzada de $5$ pliegues, ecuaci\'on \eqref{eq:vc5}, dentro de un rango posible de valores menor a $27$ (El cual surge de calcular \emph{$\sqrt{N}$}, donde \emph{N} es igual al n\'umero de observaciones usados en el entrenamiento de modelo, el cual es de $730$). De esta manera, el $n$ que  minimiza el promedio de error de  \emph{MAE, RMSE} y $R^2$ es igual a $14$, ver ecuaciones \eqref{eq:mae}, \eqref{eq:rmse} y \eqref{eq:r2}. Con este valor se procede a dividir los datos con una selecci\'on aleatoria entre entrenamiento y prueba de $70\%$ y $30\%$ respectivamente. 

\begin{figure}[H]
\centering
\subfigure[$11$ covariables] {\includegraphics[width=75mm]{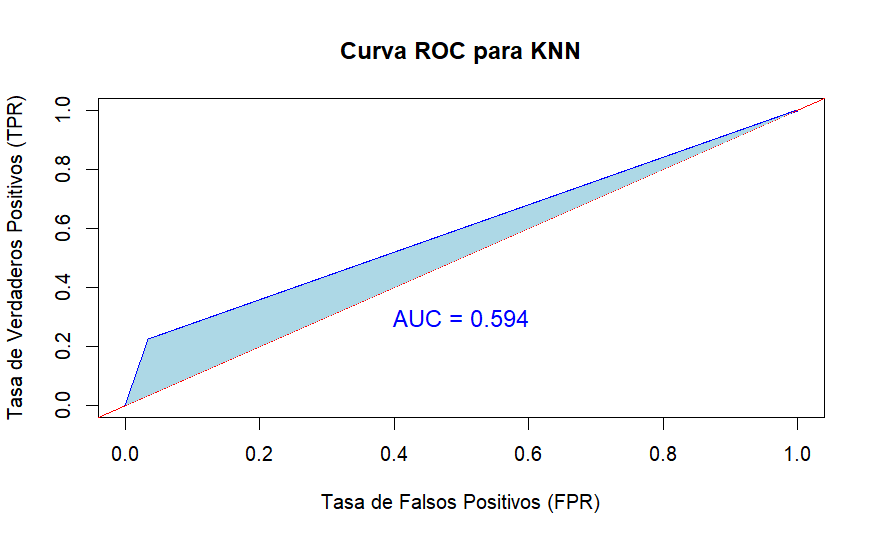} \label{fig:knn11}}
\subfigure[$6$ covariables]{\includegraphics[width=75mm]{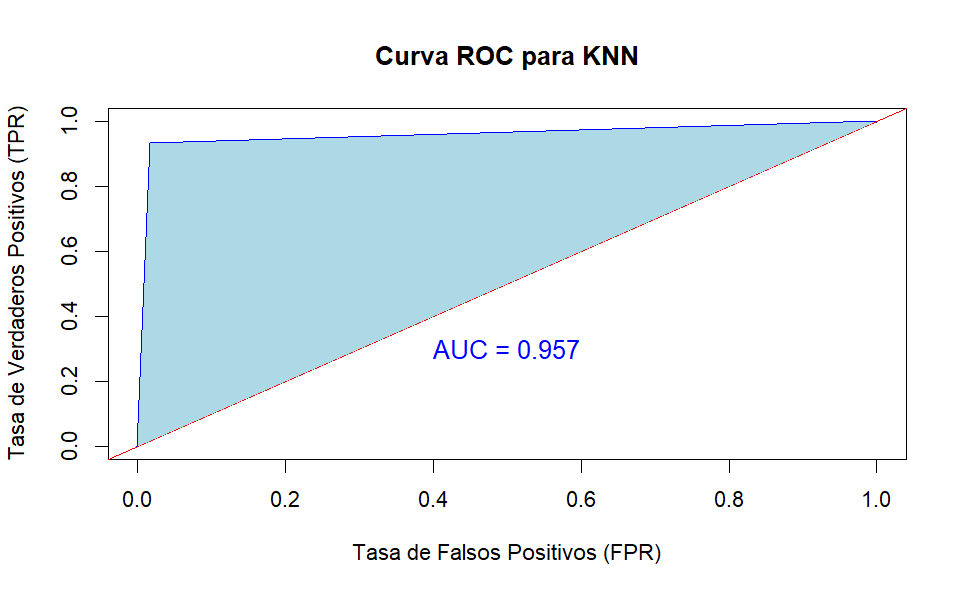}\label{fig:knn6}}
\caption{ Resultados del modelo $14$-NN para (a)~{11} y (b)~{6}, covariables.}
\label{fig:knnres}
\end{figure}

El modelo de clasificaci\'on de vecinos cercanos, optimizado con validaci\'on cruzada qu\'intuple, demostr\'o una mejora considerable en su precisi\'on, a ra\'iz de las selecciones de caracter\'isticas propuestas en el \emph{pipeline}. 
Para demostrar la potencialidad de este enfoque, consid\'erese las $11$ caracter\'isticas dadas por RF y correlaci\'on, contra las $6$ caracter\'isticas dadas por el modelo de regresi\'on log\'istica. Como se puede observar en las Figuras~\ref{fig:knn11} y \ref{fig:knn6}, se destaca la capacidad del modelo para distinguir entre clases positivas que mostr\'o un incremento importante, en particular en el caso de las clases negativas en donde la especificidad escal\'o de $0.22$ con $11$ caracter\'isticas a $0.95$ con $6$ caracter\'isticas en la identificaci\'on de ni\~nos con dificultades en el lenguaje. Lo cual sugiere que el \emph{pipeline} dise\~nado es una buena herramienta en la detecci\'on de SLI.

Para obtener una visi\'on integral de la eficacia del modelo, se utilizaron las m\'etricas de precisi\'on, F1-score, sensibilidad y especificidad. Los resultados, respaldan su utilidad cl\'inica para la detecci\'on temprana de SLI, con suficiente evidencia estad\'istica para afirmar que, los resultados no son casuales. C\'omo pueden verse en la Tabla~\ref{tab:metricas}, la combinaci\'on de m\'etodos de selecci\'on de caracter\'isticas permiti\'o, a trav\'es de las m\'etricas de predicci\'on evaluadas, determinar que el \emph{pipeline} sugerido, puede ser una buena herramienta de an\'alisis. 

\begin{table}[hb]
\caption{Resultados del Modelo $14$-NN propuesto.}
\centering
\begin{tabular}{|c|c|c|c|c|}
\hline\hline
\textbf{caracter\'isticas}&\multicolumn{4}{|c|}{\textbf{Resultados}} \\
\cline{2-5} 
\textbf{} & \textbf{\textit{Precisi\'on}} & \textbf{\textit{F1-score}} &\textbf{\textit{Sensibilidad}}& \textbf{\textit{Especificidad}} \\
\hline
11 & 82.8\% & 33.1\% & 96.4\% & 22\%  \\
\hline
6 & 97.13\% & 98.74\% & 98.71\% & 95.06\%  \\
\hline\hline
\multicolumn{4}{l}{} \\
\end{tabular}
\label{tab:metricas}
\end{table}

Adem\'as, cabe notar, que la reducci\'on de dimensionalidad fue bastante \'optima, pues se pasaron de $43$ variables iniciales a $6$ finales. Es interesante notar que las m\'etricas mejoraron ostensiblemente al usar solamente las $6$ variables propuestas. una sensibilidad del $98\%$ y una especificidad del $93\%$, indican su capacidad para identificar tanto los casos positivos como los negativos con alta precisi\'on. Por otro lado, el F1-Score, que combina precisi\'on y recall, alcanza un valor del $98\%$ y $97\%$ respectivamente. 
Estos resultados son respaldados por la matriz de confusi\'on presentada en la Tabla~\ref{tab:confusion}. La matriz se compone de los resultados del modelo en las predicciones de los valores de prueba, que componen el $30\%$ de los datos por un total de 314 observaciones. El modelo logr\'o predecir correctamente 238 casos positivos y 67 casos negativos. Sin embargo, se observaron 5 falsos negativos y 4 falsos positivos en las predicciones. Estos resultados sugieren una mejora sustancial en la precisi\'on al reducir el n\'umero de variables, con un F1-score mucho m\'as alto usando $6$ caracter\'isticas.

\begin{table}[hb]
\caption{Matriz de confusi\'on de $14$-NN con $6$ caracter\'isticas. Note que, las celdas representan el conteo de TP,TN,FP y FN de la matriz de confusi\'on del modelo $14$-NN}
    \centering
    \begin{tabular}{|c|c|c|}
    \hline\hline
    & \text{Positivo (Predicho)} & \text{Negativo (Predicho)} \\
    \hline\hline
    \text{Positivo (Real)} & 238 & 5 \\
    \hline
    \text{Negativo (Real)} & 4 & 67 \\
    \hline\hline
    \end{tabular}
    \label{tab:confusion}
\end{table}

Los resultados obtenidos sugieren que la combinaci\'on de t\'ecnicas de selecci\'on de caracter\'isticas, como \emph{Random Forest} y regresi\'on log\'istica, puede ser una herramienta potente para mejorar la eficacia del modelo de clasificaci\'on. La reducci\'on del conjunto de variables a se\'is destac\'o la relevancia de caracter\'isticas espec\'ificas, como \emph{el uso de Verbos sin declinar, la media de morfemas por oraci\'on, los errores cometidos. el promedio de silabas por palabra, la frecuencia de tipos de palabras y el uso de verbos en pasado
regular}.  En este contexto, surge la importancia de la toma de muestras de narrativas espont\'aneas de ni\~nos en el \'ambito de la psiquiatr\'ia infantil. Para ello cobran suma importancia que existan metodolog\'ias que utilicen variables cuantitativas directamente relacionadas con el desempe\~no del ni\~no, como pueden ser, el n\'umero de morfemas o el promedio de vocales por palabra. Este tipo de t\'ecnicas estandarizadas podr\'ian motivar la toma de muestras en nuevos pacientes, dado que son escasas las bases de datos y resulta complejo de acceder a material de audio o transcripciones de ni\~nos.

La presente investigaci\'on se destaca por la simpleza del modelo, su precisi\'on y su replicabilidad, en tanto no utiliza complejas variables que requieren del uso de indicadores abstractos de la psicolog\'ia o herramientas de la fonoaudiologia, que lo vuelva al experimento complejo de replicar, o hasta incluso incompatible, con transcripciones en idioma ingl\'es como espa\~nol. Es interesante notar que el \emph{pipeline} propuesto tiene la siguiente complejidad en t\'erminos de \emph{Big} $\mathcal{O}$: El m\'etodo \emph{Random Forest} $\mathcal{O}(n\log(n)Vm)$, el coeficiente de correlaci\'on de \emph{Spearman} $\mathcal{O}(n)$; regresi\'on log\'istica $\mathcal{O}(nV)$ y el clasificador de $k$ vecinos m\'as cercanos $\mathcal{O}(knV)$,  donde $V$ es el tama\~no de la caracter\'isticas y $m$ es  la profundidad de los \'arboles. Esto sugiere que el \emph{pipeline} no presenta una alta complejidad computacional.

La Tabla~\ref{tab:comparacion} muestra una comparaci\'on del \emph{pipeline} propuesto, con otros trabajos del estado-de-arte, destacando su singularidad y eficacia en t\'erminos de sus m\'etricas.  Observe que los resultados son alentadores frente a modelos de alta complejidad como las redes neuronales o SVM.

\begin{table}[H]
\caption{Comparaci\'on con algunos trabajos del estado-del arte en clasificaci\'on de SLI. DDT: Datos directos de transcripciones procesadas con t\'ecnicas de NLP, RNN: Redes neuronales recurrentes, CNN: Redes neuronales convolucionales, ANN: Redes neuronales artificiales, SVM: Support Vector Machines.}
    \centering
    \begin{tabular}{|p{5cm}|p{4cm}|p{2cm}|p{2cm}|}
        \hline\hline
         M\'etodo & Caracter\'isticas  & Precisi\'on & Ref\\
        \hline\hline
         $14$-NN & DDT & $97.77\%$ & Este trabajo \\
        \hline
         CNN & DDT & $99\%$ & \cite{Sharma2022} \\
        \hline
         SLINet & CNN 2D & $99.09\%$ & \cite{bowen1998} \\
        \hline
       likelihood ratios & Pruebas de repetici\'on de d\'igitos, no-palabras y oraciones & $94\%$ & \cite{ArmonLotem2016} \\
        \hline
      SVM, RF y RNN & Se\~nales de voz & $99.00\%$ & \cite{slogrove2020} \\
        \hline
         SVM y feed-forward neural network & Fuente glotal y coeficientes cepstrales de frecuencia \emph{Mel} & $98.82\%$ & \cite{reddy2020} \\
        \hline
         Naive Bayes, SVM y ANN & Longitud media de emisiones y estructuras gramaticales & $79\%$,$80\%$ y $76\%$ & \cite{oliva2013} \\
        \hline\hline
    \end{tabular}
    \label{tab:comparacion}
\end{table}

\section{Conclusiones}
\label{sec:con}
Este estudio se propuso un \emph{pipeline} en cascada de $3$ etapas, que permite, a trav\'es de un enfoque simple y eficiente, la detecci\'on de SLI en ni\~nos. La precisi\'on del modelo de $97.13\%$, sugiere su viabilidad cl\'inica como herramienta de detecci\'on temprana de SLI. En la primera etapa, se realiz\'o una extracci\'on de caracter\'isticas y una reducci\'on de dimensionalidad de los datos usando los m\'etodos de \emph{Random Forest} (RF) y correlaci\'on en conjunto, logrando reducir de $43$ a $11$ variables. En la segunda etapa, se estimaron las variables m\'as predictivas usando regresi\'on log\'istica, obteni\'endose $6$ variables finales de las $11$ de la etapa anterior. Estas variables son usadas en la \'ultima etapa,  para detectar el trastorno SLI en ni\~nos a partir de transcripciones de narrativas espont\'aneas. En resumen, el \emph{pipeline} dise\~nado permiti\'o reducir de $43$ variables a $6$ variables, lo que da una luz en la detecci\'on de SLI. 

El \emph{pipeline} propuesto, presenta tres fortalezas notables. Es de baja complejidad computacional; presenta una reducci\'on de dimensionalidad de los datos siguiendo criterios precisos a partir de los datos y permite un tratamiento de la informaci\'on en varios estadios, permitiendo realizar una selecci\'on de las caracter\'isticas que logran un gran nivel predictivo de la variable objetivo. La combinaci\'on de NLP y ML abre nuevas posibilidades para diagn\'osticos precisos y eficientes, con un potencial impacto en la identificaci\'on temprana y el dise\~no de intervenciones personalizadas. En cuanto a las limitaciones, se considera lo experimental del proyecto, en tanto no fue probado con otros tipos de datos; tambi\'en la estimaci\'on del valor $k$ del clasificador, puede ser \'optimo o no, en muchos casos es un valor emp\'irico que busca minimizar el error en la etapa de clasificaci\'on, por ende puede tener un rango de posibles valores. Resulta importante destacar en cuanto a las iteraciones de las regresiones que no siempre es conveniente proceder hasta hallar el m\'inimo conjunto posible de características. Dado que pueden ocurrir dos escenario. Uno donde se tiene un numero importante de variables que afectan en gran medida la capacidad del modelo de predecir la variable objetivo, y otro donde se tiene una selecci\'on reducida de características. En el primer caso se recomienda continuar efectuando regresiones con el fin de reducir la dimensi\'on de trabajo. No obstante en el segundo caso, se puede optar por no correr otra regresi\'on, en tanto se perder\'ian variables de interés por una mejoría despreciable en el modelo. Esto queda sujeto al caso de uso y la consulta de un profesional en el \'area.
Futuras investigaciones se centrar\'an en explorar la aplicaci\'on de este enfoque en poblaciones m\'as amplias, evaluar su utilidad en entornos cl\'inicos, adaptar el \emph{pipeline} a varios tipos de datos, as\'i como implementar esta metodolog\'ia en otros campos. 

\section{Disponibilidad del software}
\label{sec:soft}
El software utilizado en este estudio está disponible en la 
plataforma de desarrollo colaborativo
\url{https://github.com/SantiagoarenaDS/Pipeline-SLI-JAIIO2024}, accessed: 2024-24-03


\end{document}